# Solving the Problem of the K Parameter in the KNN Classifier Using an Ensemble Learning Approach

Ahmad Basheer Hassanat[1], Mohammad Ali Abbadi[2],
Ghada Awad Altarawneh[3]
[1,2]IT Department, [3]Accounting department
Mu'tah University
Mu'tah – Karak, Jordan

Ahmad Ali Alhasanat
College of Business Administration & Economics
Al-Hussein Bin Talal University,
Maan, Jordan

*Abstract*— **This paper presents a new solution for choosing the K parameter in the k-nearest neighbor (KNN) algorithm, the solution depending on the idea of ensemble learning, in which a weak KNN classifier is used each time with a different K, starting from one to the square root of the size of the training set. The results of the weak classifiers are combined using the weighted sum rule. The proposed solution was tested and compared to other solutions using a group of experiments in real life problems. The experimental results show that the proposed classifier outperforms the traditional KNN classifier that uses a different number of neighbors, is competitive with other classifiers, and is a promising classifier with strong potential for a wide range of applications.**

I. INTRODUCTION

The nearest neighbor approach was first introduced by [1] and later studied by [2]. This approach is one of the simplest and oldest methods used for pattern classification. It often yields efficient performance and, in certain cases, its accuracy is greater than state-of the-art classifiers [3] [4].

The KNN classifier categorizes an unlabelled test example using the label of the majority of examples among its k-nearest (most similar) neighbors in the training set. The similarity depends on a specific distance metric, therefore, the performance of the classifier depends significantly on the distance metric used [5].

The KNN classifier is one of the most popular neighborhood classifiers in pattern recognition [6] and [7], because the technique is very simple, and highly efficient in the field of pattern recognition, machine learning, text categorization, data mining, object recognition, etc. [8] and [9]. However, it has limitations, such as memory requirement and time complexity, because it is fully dependent on every example in the training set.

There are two major problems inherited from the design of the KNN [10] and [7]:

1. There is no output trained model to be used; the algorithm has to use all the training examples on each test, therefore its time complexity is linear O(*n*).

2. Its classification performance depends on choosing the optimal number of neighbors (*k*), which is different from one data sample to another.

Many studies have attempted to solve the first problem, dependent on reducing the size of the training set [11], [12], [4], [13] and [14]. Hart proposed a simple local search method called the "Condensed Nearest Neighbor" (CNN) which attempts to minimize the number of stored examples and stores only a subset of the training set to be used for classification later. Their idea is based on removing the similar redundant examples [11].

Gate presented the "Reduced Nearest Neighbor" (RNN) method, which is basically based on the CNN. The aim of the method is to further shrink the CNN stored subset by removing all examples from the subset that do not affect the accuracy of the classifier, i.e. removing them causes no significant error overall [12].

Other studies in the same vein include [15], [16], [17], [18] and [19]. Other works used some hashing techniques to increase classification speed; this includes the work of [20] and [21].

On the other hand, to the best of the authors' knowledge, there has been little work in the literature focuses on the second problem; therefore, the purpose of this study is to solve the second problem of the KNN classifier, by removing the need for using a specific *k* with the classifier.

II. RELATED WORK

Usually, the K parameter in the KNN classifier is chosen empirically. Depending on each problem, different numbers of nearest neighbors are tried, and the parameter with the best performance (accuracy) is chosen to define the classifier.

Choosing the optimal *K* is almost impossible for a variety of problems [22], as the performance of a KNN classifier varies significantly when K is changed as well as the change of distance metric used. However, it is shown in the literature that when the examples are not uniformly distributed, determining the value of K in advance becomes difficult [23].

Guo *et al*. converted the training set to another smaller domain called the "KNN Model". Their model groups each number of similar examples from the data set, based on their similarity to each other. The output model consists of tuples containing the class of the group, the similarity of the most distance point inside the group (local region) to the central data point, in addition to the number of the points of that group (region). There is no need to choose the best k, because the number of points in each group can be seen as an optimal k, i.e. different parameters are used in each group. This work is tested using six data sets, obtaining good results. Their work reduces the size of the training data, and removes the need for choosing the k parameter [10]. However, there is still a need to define other thresholds such as "error tolerant degree" and the minimum number of points allowed in each group.

Song *et al.* presented two approaches – (local informative-KNN (LI-KNN) and global informative-KNN (GI-KNN)) – to solve the problem of the k parameter in the KNN classifier. Their goal was to improve the performance of the KNN. They used a new concept, which they called "*Informativeness*". This was used as a query-based distance metric. Their experiments (based on 10 data sets from the benchmark corpus [24]) showed that their methods were less sensitive to the change of parameters than the conventional KNN classifier [22].

Hamamoto *et al.* used a bootstrap method for nearest neighbor classifier. Their experimental results showed that the nearest neighbor classifier based on the bootstrap samples outperforms the conventional KNN classifiers, mainly when the tested examples are in high dimensions [3].

Yang and Liu argue that the performance of the KNN classifier is relatively stable when choosing a large number of neighbors. They used large values for the *k* parameter such as (30, 45 and 60), and the best results of the classifier were included in their results tables [25] and [26].

Enas and Choi show that the best choice of the *k* parameter was found to be dependent on several factors, namely, the dimension of the sample space, the size of the space, the covariance structure, as well as the sample proportions [27].

The "inverted indexes of neighbors classifier" (IINC) [28], [29] and [30] is one of the best attempts found in the literature to solve the problem. The aim of their work was not intentionally to solve the problem of the k parameters; rather it was designed to increase the accuracy of the classifier. The main idea of the IINC is to use all the neighbors in the training set, rewarding the nearest neighbors, and penalizing the furthest one.

Their algorithm works as follows: the similarity distance of the test point is calculated with all the points in the training set. The distances are sorted in ascending order, keeping track of their classes. The summation of the inverted indexes is then calculated for each class using Eq(1). The probability of each class is then calculated using Eq(2). Obviously, the class with the highest probability is then predicted.

**Remark 1**: *the previous approach is based on the hypothesis that the influence of the nearest neighbors is larger than those of the furthest distance from the query point* [2], [28], [29] and [30].

The summation of the inverted indexes for class c is:

$$S_c = \sum_{i=1(c)}^{L_c} \frac{1}{i} \qquad (1)$$

where $L_c$ is the number of points of class *c*, *i* is the order of the point in the training set after sorting the distances.

The probability of a test point x belongs to a class c can be estimated as:

$$P(x|c) = \frac{S_c}{S} \qquad (2)$$

where $S = \sum_{i=1}^{N} \frac{1}{i}$

and N is the number of examples in the training set.

Jirina and Jirina argue that the experimental results based on 24 data sets taken from the benchmark corpus [24], showed that (in most tasks) the IINC outperformed some other well known classifiers such as the traditional KNN, support vector machines, decision trees, artificial neural networks, and naive Bayes classifiers. Therefore there can be an alternative to standard classification methods [28], [29] and [30].

III. THE PROPOSED WORK

There are three problems associated with the reported IINC:

1. It requires all the points in the training data to be used to calculate all the inverted indices; this prevents any attempt to reduce the size of the training set and enforces time consuming.

2. There is bias against the class of the smallest number of points; even if some of those points are around the query point, still the points far away from the query point somehow contribute to increase the probability of the class of the largest number of points. Even if each single contribution of each point get smaller as the points go further, when adding together with large number of points (examples) the contribution become significant.

3. Distances need to be sorted in ascending order to calculate the inverted indices; this take as long a time, at least O(*nlogn*) if quick sort is used; this is worse than the traditional KNN algorithm, which takes a linear time.

We propose to use ensemble learning using the same nearest neighbor rule. Basically, the traditional KNN classifier is used each time with a different K. Starting from k=1 to k = the square root of the training set, each classifier votes for a specific class. Then our multi classifiers system uses majority rule to identify the class, i.e. the class with the highest number of votes (by 1-NN, 3-NN, 5-NN… √n-NN) is chosen.

We choose to have a maximum number of classifiers to be not greater than the square root of the training data set size, because the often used rule of thumb is that k equals the square root of the number of points in the training data set [28], [29] and [30]. Another reason is that more classifiers increases computation time. This complies with what the pilot study suggests, since using this threshold was based on benefit cost, the highest accuracy with the lowest computation time.

The proposed multi classifiers system uses the odd numbers for the k parameter for three reasons: 1) to increase the speed of the algorithm by avoiding the even classifiers; 2) to avoid the chance of two different classes having the same number of votes; and 3) the pilot experiments having the even *ks* show no significant change of the results.

Recalling remark (1), the proposed classifier gives higher weights to the decision of classifiers with the nearest neighbors. The weighted sum rule is used to combine the KNN classifiers. Empirically, we found the best weighting function is using the inverted logarithmic function as in Eq(3). Figure 1 illustrates the function used.

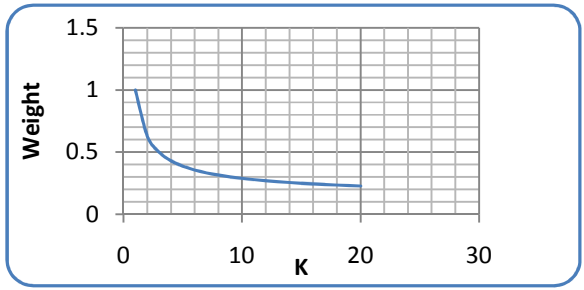

Figure 1.  Inverted logarithmic function as weighting function

$$w(k) = \frac{1}{\log_2(1+k)} \quad (3)$$

When a test example is compared with all examples in the training set, using a distance function, an array (A) is created to contain the nearest √n classes, and the weighted sum (WS) rule is defined for each class using:

$$WS_c = \sum_{k=1}^{\sqrt{n}} \sum_{i=1}^{k} \begin{cases} w(i), & A_i = c \\ 0, & otherwise \end{cases}, k = k + 2 \quad (4)$$

where for each class, we have the outer sum representing the KNN classifier for each odd k, and the inner sum calculates the weights for each classifier.

By applying Eq(4), the highest the votes for a class the highest its WS, and the nearest an example (belonging to a class) to the test example the highest its WS will be. Therefore, the predicted class is the one with the maximum weighted sum:

$$class = \mathrm{argmax}_c WS_c \quad (5)$$

To illustrate the proposed classifier, assume that we have 25 points in 2 dimensional feature space belonging to 2 different classes, in addition to one test point (the green triangle) as shown in the upper section of Figure 2.

As shown in Figure 2, the ensemble system uses the 1-NN, 3-NN and 5-NN classifiers using the weighted sum rule to find the class of the unknown point the (green triangle), which in this example is predicted to be class 1 (red square).

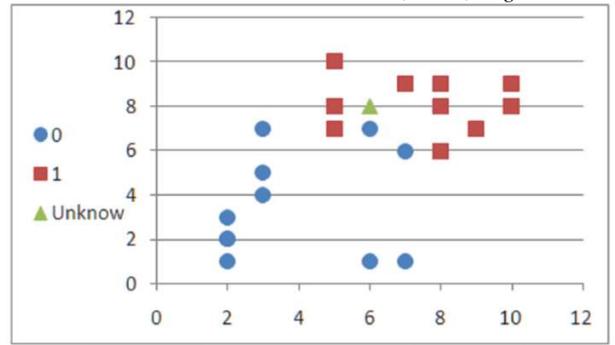

Figure 2.  Simple example showing the proposed classifier

---

**Algorithm 1:** *The proposed ensemble KNN classifier*

---

**Input**: training data set TD, test example TE

**Output**: class's index

1. **Array** Distances[n=Size(TD)]
2. index=0
3. **For each** example as E in TD {
4. Distances[index]=distanc(E,TE)//*any distance*
                                // *function*
5. index=index+1
6. }
7. Array minClasses[√n]
8. minClasses = classes (min √n Distances) //*ordered by*
                                // *distance*
9. **Array** SW[number of classes in TD]// *weight sum for*
                                // *each class*
10. Initililze SW// *fill with zeros*
11. **for** k=1 **to** √n , k=k+2
12.    **for** i=1 **to** k , i=i+1
13.      SW[minClasses[i]]=classes[minClasses[i]]+1/Log(1+i,2)
14. **return** argmax(classes)

---

Based on the time complexity analysis of algorithm 1, we can state the following theorem.

**Theorem** 1: *Time complexity of the proposed ensemble KNN classifier can be approximated to linear function O(n).*

**Proof**: Obviously, lines 1 and 2 take O(*1*), lines 3,4 and 5 take O(*n*), n is the size of the training data. Line 7 consumes O(*1*).

Line 8 consumes O(*nlog*√*n*) if we iterate the distance array n times, and insert each element into a binary search tree bounded with size √n, and remove the maximum number when the size of the tree exceeds √n.

Since √n<<n, O($log$ √n) can be approximate to a constant $k$, therefore line 8 consumes O($n k$).

Line 9 consumes O($1$). Line 10 consumes O($m$), where m is the number of classes in the training set, which normally is a constant. Thus it can be approximated to O($1$).

Line 11 consumes O($\sqrt{n}/2$) because it works only on the odd numbers, The nested loop in line 12 and the line inside (13) consume O($\sqrt{n}/2 * \sqrt{n}/2$) = O($n$). And the last line consumes O($1$).

This makes the total time complexity:

2O(1)+3O(n)+O(1)+O(n k)+ 2O(1)+O(√n/2)+2O(n)+O(1)≈O(n k)     (6)

We can write O($k$) ≈ O($1$), therefore:

$$O(n\ k) \approx O(n) \square$$

The time complexity of the proposed classifier O($n$ $log$ √n) ≈ O($n$) is better than that of the IINC, which is O($n$ $log$ $n$), because we use only the first √n nearest distances . However, if we worked the naïve version of finding the minimum k distances each time from n elements, it would then cost O($kn$), since k=√n, time complexity becomes O($n\sqrt{n}$). This function grows even faster than O($nlogn$).

## IV. RESULTS AND DISCUSSION

The proposed classifier is applied and compared to other methods that are found in the literature to solve the problem of the k parameter in the KNN classifier. For the experiments, we chose 28 different data sets to represent real life classification problems, taken from the UCI Machine Learning Repository [24]. Table 1 depicts the data sets used.

TABLE I. DESCRIPTION OF THE DATA SETS USED.

| Name | #E | #F | #C | data type | Min | Max |
|---|---|---|---|---|---|---|
| Heart | 270 | 25 | 2 | pos integer | 0 | 564 |
| Balance | 625 | 4 | 3 | pos integer | 1 | 5 |
| Cancer | 683 | 9 | 2 | pos integer | 0 | 9 |
| German | 1000 | 24 | 2 | pos integer | 0 | 184 |
| Liver | 345 | 6 | 2 | pos integer | 0 | 297 |
| Vehicle | 846 | 18 | 4 | pos integer | 0 | 1018 |
| Vote | 399 | 10 | 2 | pos integer | 0 | 2 |
| BCW | 699 | 10 | 2 | pos integer | 1 | 13454352 |
| Haberman | 306 | 3 | 2 | pos integer | 0 | 83 |
| Letter recognition | 20000 | 16 | 26 | pos integer | 0 | 15 |
| Wholesale | 440 | 7 | 2 | pos integer | 1 | 112151 |
| Australian | 690 | 42 | 2 | pos real | 0 | 100001 |
| Glass | 214 | 9 | 6 | pos real | 0 | 75.41 |
| Sonar | 208 | 60 | 2 | pos real | 0 | 1 |
| Wine | 178 | 13 | 3 | pos real | 0.13 | 1680 |
| EEG | 14980 | 14 | 2 | pos real | 86.67 | 715897 |
| Parkinson | 1040 | 27 | 2 | pos real | 0 | 1490 |
| Iris | 150 | 4 | 3 | pos real | 0.1 | 7.9 |
| Diabetes | 768 | 8 | 2 | real & integer | 0 | 846 |
| Monkey1 | 556 | 17 | 2 | binary | 0 | 1 |
| Ionosphere | 351 | 34 | 2 | real | -1 | 1 |
| Phoneme | 5404 | 5 | 2 | real | -1.82 | 4.38 |
| Segmen | 2310 | 19 | 7 | real | -50 | 1386.33 |
| Vowel | 528 | 10 | 11 | real | -5.21 | 5.07 |
| Wave21 | 5000 | 21 | 3 | real | -4.2 | 9.06 |
| Wave40 | 5000 | 40 | 3 | real | -3.97 | 8.82 |
| Banknote | 1372 | 4 | 2 | real | -13.77 | 17.93 |
| QSAR | 1055 | 41 | 2 | real | -5.256 | 147 |

#E: Number of examples. #F: Number of features. #C: Number of classes.

Each data set is divided into two data sets– one for training and the other for testing. 30% of the data set is used for testing, and the rest of the data is for training. Ten types of classifiers have been designed to compare their performances with the proposed classifier; these are 1-NN, 3-NN, 5-NN, 7-NN, 9-NN, √n –NN, 30-NN, 45-NN, 60-NN, and the IINC. These include the traditional KNN classifier using small, medium and large number of neighbors, in addition to the IINC classifier, which arguably bests state-of-the-art classifiers [28], [29] and [30].

Each classifier is used to classify the test samples using Manhattan distance. The 30% of data which were used as a test sample are chosen randomly and each experiment on each data set is repeated 10 times to obtain random examples for testing and training. Table 2 shows the results of the experiments. The accuracy of each classifier on each *normalized* data set is the average of 10 runs.

As can be seen from the results, there is no optimal k, as there is no specific number of neighbors that are suitable for all data sets to be used with the nearest neighbor rule. Each data set favors a specific number (k) of neighbors. This note justifies the proposed method, which attempts to use the power of each classifier, and employs it to enhance the overall performance of the proposed method.

According to the experiments, the using k = √n did not yield excellent results compared to other methods, so using k = √n as a rule of thumb is not a good choice for the KNN classifier. In addition to the use of a large number of neighbors such as k= 30, 45 and 60, does not help in increasing the accuracy of the KNN classifier as argued by [25] and [26]. They argued that the performance of the KNN becomes more stable when using large k. Perhaps that is because their reported results were based on text categorization data sets, while none of the above-mentioned data sets is related to the text categorization problem. Therefore, we cannot generalize their note to other data sets and classification problems.

*(IJCSIS) International Journal of Computer Science and Information Security,*
*Vol. 12, No. 8, August 2014*TABLE II. THE RESULTS OF THE PROPOSED CLASSIFIER COMPARED TO OTHER CLASSIFIERS– ACCURACIES ARE THE AVERAGE OF 10 RUNS

| Data set | 1-NN | 3-NN | 5-NN | 7-NN | 9-NN | √n -NN | 30-NN | 45-NN | 60-NN | IINC | Proposed |
|---|---|---|---|---|---|---|---|---|---|---|---|
| Australian | 0.8 | 0.86 | **0.87** | 0.86 | 0.86 | 0.86 | 0.85 | 0.86 | 0.86 | **0.87** | **0.87** |
| Balance | 0.8 | 0.8 | 0.83 | 0.85 | 0.86 | 0.88 | **0.88** | **0.88** | **0.88** | **0.88** | 0.86 |
| Banknote | **1** | **1** | **1** | **1** | **1** | 0.98 | 0.98 | 0.97 | 0.96 | **1** | **1** |
| BCW | **0.97** | **0.97** | 0.96 | 0.96 | 0.96 | 0.95 | 0.96 | 0.96 | 0.95 | 0.95 | 0.96 |
| Cancer | 0.96 | **0.97** | **0.97** | 0.96 | 0.97 | 0.96 | 0.96 | 0.96 | 0.95 | 0.95 | 0.96 |
| Diabetes | 0.69 | 0.72 | 0.73 | 0.74 | 0.75 | **0.76** | **0.76** | **0.76** | 0.75 | 0.74 | 0.74 |
| EEG | 0.84 | **0.85** | 0.84 | 0.84 | 0.84 | 0.76 | 0.81 | 0.8 | 0.79 | 0.84 | 0.83 |
| German | 0.69 | 0.7 | 0.72 | 0.73 | 0.73 | 0.74 | 0.71 | 0.71 | 0.7 | **0.74** | **0.74** |
| Glass | 0.65 | 0.66 | 0.66 | 0.65 | 0.64 | 0.64 | 0.6 | 0.53 | 0.42 | **0.68** | 0.67 |
| Haberman | 0.69 | 0.69 | 0.72 | 0.73 | 0.74 | **0.76** | 0.74 | 0.72 | 0.72 | 0.75 | 0.72 |
| Heart | 0.76 | 0.78 | 0.78 | 0.79 | 0.8 | 0.81 | **0.83** | 0.82 | **0.83** | 0.79 | 0.79 |
| Ionosphere | **0.9** | 0.89 | 0.89 | 0.89 | 0.87 | 0.85 | 0.84 | 0.78 | 0.75 | 0.85 | 0.89 |
| Iris | 0.94 | **0.96** | **0.96** | **0.96** | **0.96** | **0.96** | 0.95 | 0.95 | 0.88 | **0.96** | **0.96** |
| Letter-recognition | **0.95** | **0.95** | **0.95** | 0.94 | 0.94 | 0.82 | 0.91 | 0.9 | 0.87 | **0.95** | 0.94 |
| Liver | 0.61 | 0.63 | 0.65 | **0.66** | **0.66** | **0.66** | 0.64 | 0.64 | 0.64 | 0.64 | 0.64 |
| Monkey1 | 0.79 | 0.84 | 0.91 | 0.95 | **0.96** | 0.92 | 0.91 | 0.87 | 0.9 | 0.92 | 0.94 |
| Parkinson | 0.89 | 0.91 | 0.92 | 0.92 | 0.92 | 0.9 | 0.89 | 0.9 | 0.88 | **0.93** | **0.93** |
| Phoneme | **0.89** | 0.88 | 0.87 | 0.87 | 0.86 | 0.83 | 0.84 | 0.83 | 0.83 | 0.87 | 0.87 |
| QSAR | 0.82 | 0.84 | 0.86 | 0.85 | 0.85 | 0.84 | 0.82 | 0.81 | 0.81 | **0.86** | **0.86** |
| Segmen | **0.97** | **0.97** | 0.96 | 0.96 | 0.96 | 0.91 | 0.92 | 0.91 | 0.89 | 0.96 | 0.96 |
| Sonar | **0.87** | 0.83 | 0.81 | 0.78 | 0.75 | 0.73 | 0.75 | 0.72 | 0.71 | 0.86 | 0.85 |
| Vehicle | 0.67 | 0.67 | 0.66 | 0.66 | **0.68** | 0.66 | 0.65 | 0.62 | 0.59 | 0.67 | 0.67 |
| Vote | 0.91 | 0.93 | 0.93 | **0.94** | **0.94** | 0.93 | 0.9 | 0.89 | 0.89 | 0.93 | 0.93 |
| Vowel | **0.98** | 0.94 | 0.87 | 0.78 | 0.69 | 0.53 | 0.46 | 0.43 | 0.38 | 0.96 | 0.94 |
| Waveform21 | 0.76 | 0.79 | 0.81 | 0.82 | 0.83 | **0.85** | 0.84 | 0.84 | **0.85** | 0.83 | 0.84 |
| Waveform40 | 0.71 | 0.75 | 0.78 | 0.79 | 0.8 | **0.84** | 0.83 | 0.83 | **0.84** | 0.82 | 0.83 |
| Wholesale | 0.86 | 0.9 | **0.91** | **0.91** | **0.91** | 0.9 | 0.89 | 0.89 | 0.89 | 0.9 | **0.91** |
| Wine | **0.97** | 0.95 | 0.96 | 0.96 | 0.96 | 0.96 | 0.96 | 0.96 | 0.94 | 0.97 | 0.96 |
| *Average* | *0.83* | *0.84* | *0.85* | *0.85* | *0.85* | *0.83* | *0.82* | *0.81* | *0.8* | *0.86* | *0.86* |

On the other hand, the performance of both the proposed method and the IINC is better than the other classifiers in general. Both methods do not ask for a specific k. The good performance of the IINC is justified by the use of all the neighbors, and the good performance of the proposed method is justified by the use of ensemble learning, which makes use of weak classifiers to generate a stronger one.

It can be noted from the results that the proposed method outperformed all classifiers in 8 data sets, and even when it is behind other classifiers the difference is not more than 0.02 from the best performance. The performance of the IINC is slightly better than the proposed method, as it outperformed all classifiers in 9 data sets. However, both methods have almost the same performance in general.

It is well established in the literature [31] and according to the 'no free lunch' theorem [32], there is no optimal classifier that works perfectly for every class of problems, as the performance of the classifier depends mainly on the problem and the data used.

37

http://sites.google.com/site/ijcsis/
ISSN 1947-5500

Our method has yet another feature, which is the linear time complexity, compared to logarithmic linear time of the IINC, which needs to sort the distances to start calculating the inverted indexes. Moreover, the need for all examples in the training set prevents the IINC from speeding up using some methods such as the CNN and RNN. On the other hand, the proposed method can benefit from such methods, because it uses only the square root of the nearest neighbors.

V. CONCLUSION AND FUTURE WORK

This work proposes a new classifier based on the KNN classifier, which solves the problem of choosing the number of neighbors that participate in the final decision using the majority rule of the nearest neighbor approach. The proposed method makes use of the ensemble learning approach, where the traditional KNN is used with a different number of neighbors each time.

The experimental results using a variety of data sets of real life problems have demonstrated the superiority of the proposed method over the tradition KNN using variety of k neighbors. In addition, the proposed method was found competitive to other classifiers such as the IINC classifier. Moreover, we have shown that the speed of the proposed method (linear time) was found to be better than that of the IINC which is logarithmic linear time.

There is room for enhancing the complexity time of the proposed method using KD-trees [33] or other hashing techniques [20] and [21]. Such efforts are best left to be done in the future.

ACKNOWLEDGMENT

All the data sets used in this paper were taken from the UCI Irvine Machine Learning Repository [24], therefore the authors would like to thank and acknowledge the people behind this great corpus. Also the authors would like to thank the anonymous reviewers of this paper.

AUTHORS PROFILE

Ahmad B. A Hassanat was born and grew up in Jordan, received his Ph.D. in Computer Science from the University of Buckingham at Buckingham, UK in 2010, and B.S. and M.S. degrees in Computer Science from Mutah University/Jordan and Al al-Bayt University/Jordan in 1995 and 2004, respectively. He has been a faculty member of Information Technology department at Mutah University since 2010. His main interests include computer vision, Machine learning and pattern recognition.

Mohammad Ali Abbadi received his Ph.D and M.S. in computer science from George Washington University, USA, in 2000, and 1996 respectivly, and B.S. in computer science in 1990 from Mutah University. He has been a faculty member of Information Technology department at Mutah University since 2000. His research interests include Data Compression, Multimedia Databases & Digital Libraries, Audio/Image/Video Processing, Operating Systems, Fault Tolerance and Watermarking.

Ahmad Ali Alhasanat received his M.S. degree in Management Information Systems from Al-Balqa' Applied University/Jordan in 2014, and B.S. degree in Computer Science from Mutah University/Jordan in 2003. He has been a computer lab supervisor in College of Business Administration & Economics at Al-Hussein Bin Talal University since 2004. His main interests include management information systems and artficial intelligence.

Ghada Awad Altarawneh received her Ph.D. in Accounting from the University of Buckingham at Buckingham, UK in 2011, and B.S. and M.S. degrees in Accounting from Mutah University/Jordan and Al al-Bayt University/Jordan in 2002 and 2005, respectively. She has been a faculty member of Accounting department at Mutah University since 2011. Her main interests include Accounting information systems, and using artificial intelligence in accounting systems.